\documentclass[11pt,a4paper]{article}
\usepackage[hyperref]{acl2020}
\usepackage{times}
\usepackage{latexsym}
\usepackage{graphicx}
\usepackage{multirow}

\usepackage{microtype}

\aclfinalcopy 

\title{Annotation and Classification of Evidence and Reasoning Revisions in Argumentative Writing}

\author{Tazin Afrin, Elaine Wang, Diane Litman, Lindsay C. Matsumura, Richard Correnti\\
  Learning Research and Development Center\\
  University of Pittsburgh \\
  Pittsburgh, Pennsylvania 15260 \\
  \texttt{tazinafrin@cs.pitt.edu}, \texttt{ewang@rand.org} \\ \texttt{\{dlitman,lclare,rcorrent\}@pitt.edu}

  }

\date{}

\begin{document}
\maketitle
\begin{abstract}
Automated writing evaluation systems can improve students' writing insofar as students attend to the feedback provided and revise their essay drafts in ways aligned with such feedback. 
Existing research on revision of argumentative writing in such systems, however, has focused on the types of revisions students make (e.g., surface vs. content) rather than the extent to which  revisions actually respond to the feedback provided and improve the essay. 
We introduce an annotation scheme to capture the nature of sentence-level revisions of evidence use and reasoning (the `RER' scheme) and apply it to 5th- and 6th-grade students' argumentative essays. We show that reliable manual annotation can be achieved
and that revision annotations 
correlate with a holistic assessment of essay improvement in line with the feedback provided. Furthermore, we explore 
the feasibility of automatically classifying revisions according to our scheme.
\end{abstract}

\section{Introduction}
\label{introduction}

Automated writing evaluation (AWE) systems are intended to help improve students' writing by providing formative feedback to guide students' essay revision. Such systems are only effective if students attend to the feedback provided and revise their essays in ways aligned with such feedback. 

To date, few AWE systems assess (and are assessed on) the extent to which students' revisions respond to the feedback provided and thus improve the essay in suggested ways. Moreover, we know little about what students do when they do not revise in expected ways. 
For example, most natural language processing (NLP) work on 
writing revision focuses only on annotating and classifying  
revision purposes~\cite{daxenbergerG13g,zhang2017hh}, 
rather than on assessing the quality of a revision  in achieving its purpose.
A few studies do focus on revision quality, but without relating  revisions to  feedback~\cite{tan2014l,afrin2018improvement}.

In this study, we take a step towards advancing automated revision analysis capabilities. First, we develop a sentence-level revision scheme to annotate the nature of students' revision of evidence use and reasoning (hereafter, we refer to this as the `RER scheme') in a text-based argumentative essay writing task. By evidence use, we refer to the selection of relevant and specific details from a source text to support an argument. By reasoning, we mean an explanation connecting the text evidence to the claim and overall argument. Table~\ref{table: rev alignment and examples} shows examples of evidence and reasoning revisions from first draft to second draft. Next, we demonstrate inter-rater reliability among humans in the use of the RER scheme. In addition, we show that only desirable revision categories in the scheme relate to a holistic assessment of essay improvement in line with the feedback provided. Finally, we adapt word to vector representation features 
to automatically classify desirable versus undesirable evidence revisions, and examine how  automatically predicted evidence revisions relate to the holistic assessment of essay improvement.

%
%
%
\section{Related Work}
\label{section:related work}
Automated revision detection work has  centered on classifying edits on largely non-content level features of writing, such as spelling and morphosyntactic revisions~\cite{wisniewski2010mining}, error correction, paraphrase or vandalism detection ~\cite{daxenbergerG13g}, factual versus fluency edits~\cite{bronner2012m}, and document- versus word-level revisions ~\cite{roscoe2015automated}. Other research 
has focused on patterns of revision behavior, for example, the addition, deletion, substitution, and reorganization of information ~\cite{zhang2020engaging}. 
However, these categories  center on   general writing features and behaviors. In the context of AWE systems, this could be seen as a limitation because feedback is most useful to students and teachers alike when it is keyed to critical features of a genre – such as claims, reasons, and evidence use in argumentative writing -- that are most challenging to teach and learn. 

Some research has begun to take up the challenge of investigating student revision for argumentative 
writing~\cite{zhang2015l,zhang2017hh}. Results  show a high level of agreement for human annotation and some relationship to essay improvement, though not at the level of individual argument elements~\cite{zhang2015l}. Existing schemes also lack in specificity, e.g., they do not distinguish between desirable and undesirable  revisions for each argument element in terms of improving essay quality.



Prior work on assessing revision quality  has evaluated revision in general terms (e.g., strength~\cite{tan2014l} or overall improvement~\cite{afrin2018improvement}), but without consideration of the feedback students were provided. 
We instead focus on analyzing revisions in response to feedback from an AWE system. Although prior studies have focused on all revision categories (e.g., claim, evidence, and word-usage~\cite{zhang2015l}),  we  focus on only evidence and reasoning revisions that correspond to the scope of the AWE system's feedback. Also, we focus not only on why the student made a revision (e.g., add evidence) but also analyze if the revision was desirable or not (e.g., relevant versus irrelevant evidence).

\section{Corpus}
\label{section:corpus}
Our corpus consists of the first draft (Draft1) and second draft (Draft2) of 143 argumentative essays. The corpus draws from our effort to develop an automated writing evaluation system - eRevise, to provide 5th- and 6th-grade students feedback on a response-to-text essay~\cite{zhang2019erevise,wang2020eRevise}. 
The writing task administration involved teachers reading aloud a text while students followed along with their copy. 
Then, students were given a writing prompt\footnote{``Based on the article, did the author provide a convincing argument that winning the fight against poverty is achievable in our lifetime? Explain why or why not with 3-4 examples from the text to support your answer."} to write an argumentative essay. 

Each student wrote Draft1 and submitted their essay to the AWE system. Students then received feedback focused specifically on the use of text evidence and reasoning. Table~\ref{table:feedback messages} shows the top-level feedback messages\footnote{See ~\cite{zhang2019erevise} for detailed feedback messages.} that the system provided. Finally, students were  directed to revise their essay in response to the feedback, yielding Draft2. 

As part of a prior exploration of students' implementation of the system's feedback, this corpus of 143 essays was coded holistically on a scale from 0 to 3 for the extent to which use of evidence and reasoning improved from Draft1 to Draft 2 in line with the feedback provided ~\cite{wang2020eRevise}\footnote{In the prior study, 
two researchers double-coded 35 of the 143 essays (24 percent). 
Cohen’s kappa was 0.77, indicating ‘substantial’
agreement~\cite{mchugh2012kappa}.}.  A code, or score, of 0 indicated no attempt to implement the feedback given; 1= no perceived improvement in evidence use or reasoning, 2= slight improvement; and 3= substantive improvement. Note again that this score represents a subjective, holistic (i.e., not sentence-level) assessment of whether Draft2 improved in evidence use and/or reasoning specifically in alignment with the feedback that a particular student received. We refer to this as {\bf `improvement score'} in the rest of the paper. 

\begin{table}
\centering
\begin{tabular}{|p{0.05\columnwidth}|p{0.8\columnwidth}|}

\hline	
No	&	Feedback Message	\\ \hline
1	&	Use more evidence from the article	\\ \hline
2	&	Provide more details for each piece of evidence you use	\\ \hline
3	&	Explain the evidence	\\ \hline
4	&	Explain how the evidence connects to the main idea \& elaborate	\\ \hline
\end{tabular}
\caption{Top-level feedback from the AWE system.}
\label{table:feedback messages}
\end{table}

\subsection{Preparing the corpus for annotation}
\label{Preparing the corpus for annotation}
On average, Draft1 essays contain 14 sentences and 253 words, and Draft2 essays contain 18 sentences and 334 words. 
To prepare the corpus for annotation, we first segmented each Draft1 and Draft2 essay into sentences, then manually aligned them at sentence-level. For example, if a sentence is added to Draft2, it is aligned with a null sentence in Draft1. If a sentence is deleted from Draft1, it is aligned with a null sentence in Draft2. A modified sentence, or a sentence with no change in Draft2, is aligned with the corresponding sentence in Draft1\footnote{Sentence order substitution is evaluated as deleted then inserted.}. Based on this alignment, we then extracted the 1475 sentence pairs where students made either additions, deletions, or modifications as {\it revisions.} The remaining 1362 aligned sentences had no changes between drafts and were thus not extracted as revisions.

Each revision was next manually annotated\footnote{Annotator Cohen's kappa of 0.753.}
for its revision purpose according to the scheme proposed in~\cite{zhang2015l}, which  categorizes revisions into surface versus content changes. Surface revisions are changes to fluency or word choice, convention or grammar, and organization. Content revisions are meaningful textual changes such as claim or thesis, evidence, reasoning, counter-arguments etc. From among these revisions, only evidence and reasoning revisions are used for the current study, due to their alignment with the AWE feedback messages in Table~\ref{table:feedback messages}.

\begin{table}
\centering
\begin{tabular}{|l|c|c|c|}
\hline	
\#Sentence		&	Draft2	&	\#No Change	&	\#Revision	\\ \hline
Total		&	2652	&	1362	&	1475	\\ \hline
Avg.	&	18.545	&	9.524	&	10.315	\\ \hline



\end{tabular}
\caption{Essay statistics (N=143).}
\label{table:data statistics}
\end{table}
Table~\ref{table:data statistics} shows the descriptive statistics of the essay corpus at the sentence-level. The second column shows the total and average number of sentences for Draft2. The third column shows that, on average, about 9 sentences per essay were unchanged. The final column shows that, on average, 10 sentences per essay were revised\footnote{\#Revision also includes deleted sentences, hence \#Revision + \#No Change does not equal  \#Sentence in Draft2.}. Out of those 10 sentences, only two to three sentences were revised with respect to evidence, and another two to three sentences with respect to reasoning, on average over all 143 students. This indicates that students engaged in very limited revisions of evidence and reasoning, even when provided feedback targeted to these argument elements. 

Table~\ref{table:revision statistics} shows the statistics for the students who did revise their essay.
Note that out of 143 students, 50 students (35\%) did not make any evidence-use  revisions; 32 students (22\%) did not make any reasoning revisions. Only 10 students (7\%) did not make any evidence or reasoning revisions. 4 students (3\%) did not make any revision at all.  From these students we extracted 386 evidence revisions and 389 reasoning revisions, a total of 775 sentence-level revisions. We do not consider the other 700 revisions (claim, word-usage, grammar mistakes, etc.) in this study. 

To better understand how students did revise, whether their revisions were desirable, and whether desirable revisions relate to a measure of essay improvement that includes alignment with feedback, we developed a revision categorization scheme and conducted the analysis described below.

\begin{table}
\centering
\begin{tabular}{|l|c|c|c|}
\hline
	&	Evidence	&	Reasoning	&	Other	\\
\#Revision	&	(N=93)	&	(N=111)	&	(N=129)	\\ \hline
Total	&	386	&	389	&	700	\\ \hline
Min	&	0	&	0	&	0	\\ \hline
Max	&	36	&	17	&	21	\\ \hline
Avg.	&	4.151	&	3.505	&	5.426	\\ \hline
\end{tabular}
\caption{Revision statistics.}
\label{table:revision statistics}
\end{table}

\section{Revision Categorization (RER Scheme)}
\begin{table*}
\centering
\begin{tabular}{|p{0.55\columnwidth}|p{0.6\columnwidth}|p{0.2\columnwidth}|p{0.2\columnwidth}|p{0.2\columnwidth}|}
\hline
\textbf{Draft1}	&	\textbf{Draft2}	&	\textbf{Operation}	&	\textbf{Purpose}	&	\textbf{RER code}	\\ \hline
In the story, ``A Brighter Future," the author convinced me that ``winning the fight against poverty is achievable in our life time."	&	In the story, ``A Brighter Future," the author of the story convinced me that winning the fight against poverty is achievable in our lifetime.	&	Modify	&	Fluency	&		\\ \hline
    &	I think that in Sauri, Kenya [where poverty is all around], people were in poverty.	&	      Add	&	Claim	&		\\ \hline
	&	In the story it states ``The Yala sub-District Hospital has medicine."	&	Add	&	Evidence	&	Relevant	\\ \hline
For example, we have good food and clean water	&		&	Delete	&	Evidence	&	Irrelevant	\\ \hline
	&	This shows that there was a change at the hospital because they had medicine which is good for the peoples health when they get sick.	&	Add	&	Reasoning	&	Linked to Claim and Evidence	\\ \hline
\end{tabular}
\caption{Example revisions from aligned drafts of an essay and application of RER codes.}
\label{table: rev alignment and examples}
\end{table*}

We propose a new scheme for annotating revisions of evidence use and reasoning (RER scheme) that will be useful for assessing the improvement of the essay in line with the feedback provided. The initial set of codes drew from the qualitative exploration of students' implementation of feedback from our AWE system~\cite{wang2020eRevise}, in which the authors inductively and holistically coded how students successfully and unsuccessfully revised their essays with respect to evidence use and reasoning. For example, students sometimes added evidence that repeated evidence they had already provided in Draft1. Or they successfully modified sentences to better link the evidence to the claim. 

Both the initial set of codes and our AWE system's feedback messages were informed by writing experts and research suggesting that strong argument writing generally features multiple pieces of specific evidence that are relevant to the argument and clear explication (or reasoning) of how the evidence connects to the claim and helps to support the argument (see, for example, ~\cite{delapaz2012adolescents,ohallaron2014argwriting,wang2018rta}).

For the present study, two annotators read through each extracted evidence or reasoning-related revision in the context of the entire essay. They labeled each  instance of revision with a code. The annotators iteratively expanded or refined the initial codes until they finalized a set of codes for evidence use revisions and another for reasoning revisions (see sections 4.1 and 4.2). Together these two sets comprise the RER scheme.
Subsequently, the two annotators applied the RER scheme to all instances of evidence use or reasoning-related revisions in all 143 students' essays.\footnote{33 of the essays, or 23 percent, were double-coded for reliability, see Section~\ref{section: evaluation} for kappa score.} 
Annotators selected the best code; no sentence received more than one code.

Table~\ref{table: rev alignment and examples} presents an example of corpus preparation (Operation and Purpose, section~\ref{Preparing the corpus for annotation}) and RER coding (see below) as applied to an excerpted  essay and its revision. 
Table~\ref{table:RER examples} presents an example of each code, though for parsimony, we only present additive revisions -- not deletions or modification, as these are less common.
Table~\ref{table:distribution} shows the distribution for each RER code.

\subsection{Revision of evidence use}
\label{subsec: evidence scheme}
Revisions related to evidence  are characterized by one of the following five codes. All codes apply to added, deleted, or modified revisions, except `Minimal', which only applies to modified evidence.
\textbf{Relevant} applies to examples or details that support (i.e., are appropriate and related to) the particular claim.
\textbf{Irrelevant} applies to examples or details that are unnecessary, impertinent to, or disconnected from the claim. They do not help with the argument.
\textbf{Repeat evidence} applies to examples or details that were already present in Draft1; students are merely repeating the information.
\textbf{Non-text based} applies to examples or details outside of the provided text.
\textbf{Minimal} applies to minor modifications to existing evidence that may add some specificity, but do not affect the argument much.

\subsection{Revision of reasoning}
\label{subsec: reasoning scheme}
Reasoning revisions are characterized by one of the following six codes. All codes apply to added, deleted, or modified revisions, except `Minimal', which only applies to modified reasoning.
\textbf{Linked claim-evidence (LCE)} applies to an explanation that connects the evidence provided with the claim.
\textbf{Not LCE} applies to an explanation that does not connect the evidence provided with the claim.
\textbf{Paraphrase evidence} applies to an attempt at explanation that merely paraphrases the evidence rather than explain or elaborate upon it.
\textbf{Generic} applies to a non-specific explanation that is reused multiple times, after each piece of evidence (e.g., ``This is why I am convinced that we can end poverty.")
\textbf{Commentary} applies to an explanation that is unrelated to the main claim or source text; most of the time, it comes from the writer's personal experience.
\textbf{Minimal} applies to minor modifications that do not affect the argument much.

\begin{table*}
\centering
\begin{tabular}{|p{0.16\textwidth}|p{0.8\textwidth}|}
\hline
	&	\textbf{Example of ``Add" Revision (``Modify" for Minimal Revision)}	\\ \hline
\textbf{Evidence}	&		\\ \hline
*Relevant	&	To support the point that conditions in Sauri were bleak, a student added this new example: ``The hospitals don't have the medicine for their sick patients so therefore they can get even more ill and eventually die [if] the [immune] system is not strong enough."	\\ \hline
Irrelevant	&	To support the claim that winning the fight against poverty is possible, the student wrote, ``Students could not attend school because they did not have enough money to pay the school fee." This does not support the claim.	\\ \hline
Repeat Evidence	&	``Malaria causes adults to get sick and cause children to die" was added as sentence \#27 in a student's Draft2, but sentence \#5 already said, ``Around 20,000 kids die a day from malaria and the adults get very ill from it."	\\ \hline
Non-Text-Based	&	Student provided example of an uncle living in poverty, rather than draw from examples in the source text about poverty in Kenya.	\\ \hline
Minimal	&	In Draft1, the student wrote, ``Now during the project there are no school fees, the schools serve the students lunch, and the attendance rate is way up." In Draft 2, the student specified “Millennium Villages” project.	\\ \hline
\textbf{Reasoning}	&		\\ \hline
*LCE	&	The student argued that we can end poverty because Sauri has already made significant progress. After presenting the evidence about villagers receiving bednets to protect against malaria, the student added, ``This shows that the people of Sauri have made progress and have taken steps to protect everyone using the bed nets and other things."	\\ \hline
Not LCE	&	The student claims that Sauri is overcoming poverty. After presenting the evidence that ``Each net costs \$5," the student wrote, ``This explain how low prices are but we may not get people to lower them more."	\\ \hline
Paraphrase	&	After presenting the evidence that ``People's crops were dying because they could not afford the necessary fertilizer," the student added, ``This evidence shows that the crops were dying and the people could not get the food that they needed because the farmers could not afford any fertilizer…"	\\ \hline
Generic	&	After the first piece of evidence, the student added, ``This evidence helps the statement that there was a lot of poverty." Then after the second piece of evidence, the student added almost the same generic sentence, ``This statement also supports that there were a lot of problems caused by poverty."	\\ \hline
Commentary 	&	After a piece of evidence, a student wrote, ``We think that we are poor because we can not get toys that we want, but we go to school and its not free."	\\ \hline
Minimal	&	In Draft1, the student wrote, ``I believe that because it states that we have enough hands and feet to get down and dirty and help these kids that are suffering.". In Draft2, the student only added ``and are in poverty" to the end of the sentence.	\\ \hline
\multicolumn{2}{p{\textwidth}}{* indicates desirable revision, as  the revision has hypothesized utility in improving the assigned essay in alignment with provided feedback given in Table 1. Other codes may also be desirable given a different writing task with different feedback (e.g., students may be asked to provide  non-text-based evidence from their own experience).}
\end{tabular}
\caption{Example of each RER code.}
\label{table:RER examples}
\end{table*}




\begin{table*}
\centering
\begin{tabular}{|l|r|r|r|r|}
\hline										
RER Code	&	Add	&	Delete	&	Modify	&	Total	\\ \hline	\hline
Evidence	&	265	&	63	&	58	&	386	\\ \hline	
\quad Relevant	&	159	&	50	&	30	&	239	\\ \hline	
\quad Irrelevant	&	26	&	9	&	8	&	43	\\ \hline	
\quad Repeat Evidence	&	70	&	4	&	0	&	74	\\ \hline	
\quad Non-Text-Based	&	10	&	0	&	0	&	10	\\ \hline	
\quad Minimal	&	0	&	0	&	20	&	20	\\ \hline	\hline
Reasoning	&	270	&	59	&	60	&	389	\\ \hline
\quad LCE	&	90	&	18	&	13	&	121	\\ \hline	
\quad Generic	&	20	&	0	&	1	&	21	\\ \hline	
\quad Paraphrase	&	50	&	10	&	5	&	65	\\ \hline	
\quad Not LCE	&	62	&	11	&	12	&	85	\\ \hline	
\quad Commentary	&	48	&	20	&	7	&	75	\\ \hline	
\quad Minimal	&	0	&	0	&	22	&	22	\\ \hline	\hline
Total	&	535	&	122	&	118	&	775	\\ \hline	
\end{tabular}
\caption{RER code distribution (N=143).}
\label{table:distribution}
\end{table*}

\section{Evaluation of the RER Scheme}
\label{section: evaluation}
We evaluated our annotated corpus to answer the following research questions (RQ):

\textbf{RQ1:} What is the 
inter-rater reliability for annotating 
revisions of evidence use and reasoning?

\textbf{RQ2:} Is the number of each type of revision related to essay `improvement score'? 

\textbf{RQ3:} Is there any difference in the  `improvement score' based on the kinds of revisions? 

\textbf{RQ4:} Is there a cumulative benefit to predicting essay `improvement score' when students made multiple types of revisions? 


To answer RQ1, we calculated Cohen's kappa for inter-rater agreement on the 33 essays (23 percent) that were double-coded. Our results show that we were able to achieve substantial inter-rater agreement on reasoning (k = .719) and excellent inter-rater agreement for evidence use (k = .833) (see, e.g., ~\cite{mchugh2012kappa}).

\begin{table*}
\centering
\begin{tabular}{|l|c|c|c|c|}
\hline	
RER Code	&	Add	&	Delete	&	Modify	&	Total	\\ \hline
\hline
Evidence	&	0.17*	&	0.00	&	0.13	&	0.15	\\ \hline	
\quad Relevant	&	0.25**	&	0.02	&	0.09	&	0.20*	\\ \hline	
\quad Irrelevant	&	0.05	&	-0.00	&	0.07	&	0.06	\\ \hline	
\quad Repeat Evidence	&	0.01	&	-0.06	&	--	&	0.00	\\ \hline	
\quad Non-Text-Based	&	0.07	&	--	&	--	&	0.07	\\ \hline	
\quad  Minimal	&	--	&	--	&	0.06	&	0.06	\\ \hline \hline
Reasoning	&	0.40**	&	0.09	&	-0.10	&	0.30**	\\ \hline	
\quad  LCE	&	0.45**	&	0.05	&	0.09	&	0.41**	\\ \hline	
\quad  Generic	&	-0.03	&	--	&	-0.04	&	-0.04	\\ \hline	
\quad  Paraphrase	&	0.22**	&	0.09	&	0.02	&	0.22**	\\ \hline	
\quad  Not LCE	&	0.09	&	0.00	&	-0.07	&	0.04	\\ \hline	
\quad Commentary	&	-0.02	&	0.08	&	-0.08	&	0.01	\\ \hline	
\quad Minimal	&	--	&	--	&	-0.14	&	-0.14	\\ \hline	


\end{tabular}
\caption{Revision correlation to `Improvement Score'. (N=143,  *  p$<.05$,  **  p$<.01$)}
\label{table:correlation}
\end{table*}

To answer RQ2, we calculated the Pearson correlation between the raw number of revisions per code to the `improvement score' described in Section~\ref{section:corpus}.  Table~\ref{table:correlation} shows that the total number of evidence-related revisions was not significantly correlated with `improvement score' (r = .15), while the total number of reasoning revisions was 
(r = .30). Table~\ref{table:correlation} also shows that positive correlations were found for added evidence or reasoning (r = .17 and .40, respectively), whereas deletions and modification were not significantly correlated.

Looking at the correlations for our proposed RER codes  (which sub-categorize the Evidence and Reasoning codes~\cite{zhang2015l}), we see that the RER codes yield more and generally stronger results.
We found that, as hypothesized, adding relevant pieces of evidence was significantly positively correlated with the `improvement score', while the addition of irrelevant evidence, non-text based evidence or repeating prior evidence were all unrelated to this score.
Similarly, we found that adding reasoning that linked evidence to the claim (LCE) was significantly correlated with the `improvement score' and so was paraphrasing evidence. Other reasoning codes, as expected, were not significantly related to the `improvement score'. We did not initially consider 
paraphrases as a desirable type of revision; yet, this code showed a  significant positive correlation.
While unexpected, we were not altogether surprised as two of the feedback messages (shown in Table~\ref{table:feedback messages}) did explicitly ask for students to put ideas into their own words (see ~\cite{zhang2019erevise} for details).
Although addition of evidence and reasoning revisions demonstrated correlation to the `improvement score', deletions and modifications did not show any intuitive correlation. We suspect that this is due to the comparatively small number of delete and modify revisions.

\begin{table}
\centering
\begin{tabular}{|l|l|l|l|}
\hline										
RER Code	&	Add		&	Modify	&	Total	\\ \hline	\hline
Evidence:	&	\multicolumn{3}{c|}{}	\\ \hline	
\quad Relevant	&	3*$>$1			&	3*$>$2	&		\\ \hline
\hline

Reasoning:	&			\multicolumn{3}{c|}{}	\\ \hline	
\quad LCE &	3*$>$0,1,2	&			3*$>$1	&	3*$>$0,1,2	\\ \hline	
\quad Not LCE	&	2*$>$0,3			&		& 2*$>$0,3	\\ \hline	

\end{tabular}
\caption{ANOVA results showing differences among `Improvement Scores' (coded as 0, 1, 2, 3). Only categories with significant results are shown. All categories were tested. (N=143, * p$<.05$,  ** p$<.01$)}
\label{table:anova}
\end{table}

To answer RQ3, we performed one-way ANOVAs 
for different levels of the `improvement score' (0=no attempt, 1=no improvement, 2=slight improvement, 3=substantive improvement, aligned with feedback provided) comparing means of the number of revisions added, modified, or deleted.  ANOVAs showed overall significance for the categories shown in Table~\ref{table:anova}. 
Tukey post-hoc analyses showed that students whose essays substantively improved made more revisions in which they added or modified relevant pieces of evidence.  Students who substantively improved also added or modified their reasoning linking evidence to their claims (LCE) more than students in all other groups.
Finally, students with slightly improved essays added more explanations not linking evidence to claim (Not-LCE) than did students who made no attempt at revision or  whose essays substantively improved.

\begin{table}
\centering
\begin{tabular}{|l|l|r|r|}
\hline										
\textbf{Model}	&	\textbf{Variables}	&	\textbf{Coef.}		&	$\textbf{R}^\textbf{2}$	\\ \hline	\hline
Model\_E	&	add Relevant	&	0.25**		&	0.06	\\ \hline	\hline
\multirow{2}{*}{Model\_R}	&	add LCE	&	0.05**		&	\multirow{2}{*}{0.25}	\\ \cline{2-3}	
	&	add Paraphrase	&	0.08**		&		\\ \hline	\hline
\multirow{4}{*}{Model\_ER}	&	add LCE	&	0.45**		&	\multirow{4}{*}{0.32}	\\ \cline{2-3}	
	&	add Paraphrase	&	0.20**		&		\\ \cline{2-3}	
	&	add Relevant	&	0.29**	&		\\ \cline{2-3}	
	&	del Relevant	&	-0.21*		& 	\\ \hline
	
\end{tabular}
\caption{Stepwise linear regression results predicting `Improvement Score'. (N=143, * p$<.05$,  ** p$<.01$)}
\label{table:lr}
\end{table}

To answer RQ4, we examined three stepwise linear regression models to understand whether adding more revision codes had a cumulative influence explaining more variance in  `improvement score'. $Model\_E$ included only revisions related to evidence use. $Model\_R$ included only revisions related to reasoning.  $Model\_ER$ included all evidence use and reasoning revisions.
As shown in Table~\ref{table:lr}, $Model\_ER$ shows significant positive coefficients for the addition of relevant evidence, reasoning that links evidence to the claim (LCE), and reasoning that paraphrases evidence. 
The positive relationship shows that more of these kinds of revisions are more likely to lead to a higher `improvement  score'. Note that the order of the coefficients is related to the magnitude of the r-squared they explain - thus linked claim and evidence (LCE) has the strongest relationship with the score. Meanwhile, deleting relevant pieces of evidence has a negative relationship when adjusting for the other covariates in the model, which means that, all else being equal, this is an undesirable revision. 
\section{Automatic RER Classification}
\label{revision classification}
The ultimate goal for developing the RER scheme is to implement it in an AWE system to provide feedback to students about revision outcome not only at the essay-level but also
at a more actionable, sentence-level. While the previous section demonstrated the utility of the RER scheme, this section explores its automatic classification.
Since our overall revision dataset is small, we focus on the simplified task of developing a binary classifier to predict whether an Evidence revision is `Relevant' or not. `Relevant' is both the most frequent RER code and relates positively to the improvement score.

The input  is a revision sentence pair -- the sentence from Draft1 (S1) and its aligned sentence from Draft2 (S2). The  pair can have 3 variations: (null, S2) for added sentences, (S1, null) for deleted sentences, and (S1, S2) for modified sentences.
Since we are focusing on `Relevant' evidence prediction, and by our definition in Section~\ref{subsec: evidence scheme} `Relevant' evidence  supports the claim, we also consider the given source text (A) in extracting features.

\textbf{Features.} We explore Word2vec as features for our classification task\footnote{We also explored n-gram features from a previous revision classification task~\cite{zhang2015l}. Our  classification algorithm performed better with word2vec features.}. We extract representations of S1, S2, and A using the pre-trained GloVe word embedding ~\cite{pennington2014glove}. For each word representation (w) we use the vector of dimension 100, $w=[v_1, \dots , v_{100}]$. Then the sentence or document vector (d) is calculated as the average of all word vectors $d = [d_1, \dots , d_{100}]$, where $d_i = mean(v_{1i}, \dots, v_{ni})$, for n words in the document. Following this method we extract vectors $d_{s1}$, $d_{s2}$, and $d_a$ for S1, S2, and A respectively. Finally, we take the average of those 3 vectors to represent the feature vector, $f = [f_1, \dots , f_{100}]$, where $f_j = mean(d_{s1j}, d_{s2j}, d_{aj})$.

For machine learning, we use off-the-shelf 
Logistic Regression (LogR) from the 
scikit-learn 
toolkit.\footnote{We also explored Support Vector Machines (SVM) but Logistic Regression outperformed  SVM in our experiment.} We did not perform any parameter tuning or feature selection. 
In an intrinsic evaluation, we compare whether there are significant differences between the classifier's performance and a majority baseline in terms of average un-weighted precision, recall and F1, using paired sample t-tests over 10-folds of cross-validation. In an extrinsic evaluation, we repeat the Pearson correlation study in Section~\ref{section: evaluation} for the predicted code, `Relevant' evidence.
\begin{table}
\centering
\begin{tabular}{|l|c|c|c|}
\hline			
	&	Precision	&	Recall	&	F1-score	\\ \hline
Majority	&	0.309	&	0.500	&	0.377	\\ \hline
LogR	&	\textbf{0.615**}	&	\textbf{0.622**}	&	\textbf{0.594**}	\\ \hline

\end{tabular}
\caption{10-fold cross-validation result for classifying Evidence as `Relevant' or not. (N=386, ** p$<.01$)}
\label{table:classification-prf}
\end{table}

\begin{table}
\centering
\begin{tabular}{|l|l|c|c|c|c|}
\hline			
		&	Add	&	Delete	&	Modify	&	Total	\\ \hline
Gold	&	0.25**	&	0.02	&	0.09	&	0.20*	\\ \hline
Majority	&	0.17*	&	0.00	&	0.13	&	0.15	\\ \hline
LogR	&	0.17*	&	-0.01	&	0.03	&	0.15	\\ \hline
\end{tabular}
\caption{Correlation of  predicted `Relevant' evidence to  `Improvement Score'. (N=143, * p$<.05$,  ** p$<.01$)}
\label{table:classification-corr}
\end{table}

\textbf{Intrinsic evaluation.} Table~\ref{table:classification-prf} presents the results of the binary classifier predicting `Relevant' evidence. The results show that the logistic regression classifier significantly outperforms the baseline using our features for all metrics.

\textbf{Extrinsic evaluation.} Table~\ref{table:classification-corr} shows the Pearson correlation of  `Relevant' evidence to `Improvement Score' using  `Gold' human labels (repeated from Table~\ref{table:correlation}) versus  predicted  labels from the majority and logistic regression  classifiers. First, the number of  `Add Relevant' revisions, whether gold or predicted,
significantly correlates to improvement.  
While it is not surprising that the correlation is lower for LogR than for  Gold (upper bound), it is unexpected that LogR and Majority (baseline) are the same.  This likely reflects the  Table~\ref{table:correlation} result that adding any type of Evidence, relevant or not, correlates with improvement. 
In contrast, the predicted models are not yet accurate enough to replicate the statistical significance of the `Gold' correlation between improvement and `Total Relevant' revisions.
%
%
%

\section{Discussion}
In our corpus, students revised only about half of the sentences from Draft1 to Draft2. Among the revisions, only a small proportion focused on evidence or reasoning, despite 
feedback 
targeting these argument elements exclusively. This resonates with writing research (though not in the context of AWE) showing that students often struggle to revise~\cite{faigley1981w,macarthur2018evaluation}, and that novice writers  -- like our 5th- and 6th-graders -- 
tend to focus on local word- and sentence level problems 
rather than content or structure ~\cite{macarthur2004,macarthur2018evaluation}. When novices do revise, their efforts frequently result in no improvement or improvement only in 
surface features
~\cite{patthey2004urbanschool}.

We knew of no revision schemes that assessed the extent to which evidence use and reasoning-related revisions aligned with desirable features of argumentative writing (i.e., showed responsiveness to  system feedback  to use more relevant evidence, give more specific details, or provide explanations connecting evidence to the claim); hence, we developed the RER scheme. The scheme -- along with the reliability we established and the positive correlations we demonstrated between its sentence-level application and a holistic assessment of essay improvement in line with provided feedback -- is an important contribution because 
the codes are keyed to critical features of the argument writing genre. Therefore, it is more useful than existing schemes that focus on general revision purposes (surface vs. content) or operations (addition, deletion, modification) for assessing the quality of students' revisions.

This assessment capability is important for at least two related reasons. First, an AWE system is arguably only effective if it helps to improve  writing in line with any feedback provided. It is easier to attribute other types of revisions or improvements to the general opportunity to revisit the essay than to any inputs the system provides to students. 
For argument writing (and our AWE system), then, it is necessary, to be able to identify specific revision behaviors related to evidence use and reasoning. 
With the RER scheme, we were able to distinguish among revision behaviors. On the whole, predictably undesirable revisions (e.g., deleting relevant evidence) were not correlated with the `improvement score'.

Second, gaining insight into how  students specifically revise evidence use and reasoning can help hone the content of AWE feedback so that it better supports students to make desirable revisions that impact the overall argument quality. From our coding, we learned that students make deletion or modification revisions less frequently; rather they tend to make additions, even if they do not improve the essay. We also learned that repeating existing evidence accounted for about 19 percent of the evidence-use revisions. We could refine our feedback  to preempt students from making these undesirable revisions. Or, once automated revision detection is implemented, we could develop a finer-grained set of feedback messages to provide students to guide their second revision (i.e., production of Draft 3).


Finally, our study takes a step towards advancing automated revision detection for AWE by developing a simple machine learning algorithm for classifying relevant evidence. However, it is important to note that the classifier's input is currently based on the gold (i.e., human) alignments of  the essay drafts and the gold revision purpose labels (e.g., Evidence).  An actual end-to-end system would have lower performance  due to the propagation of errors from both alignment and revision purpose classification. In addition, due to the small size of our current corpus, our classification study was simplified to
focus on evidence rather than both evidence and reasoning, and to focus on binary rather than 5-way classification.
Although our algorithm is thus limited to predicting only relevant evidence, the classifier nonetheless outperforms the baseline given little training data. 


\section{Conclusion and Future Work}

We developed the RER scheme as a step towards advancing automated revision detection capabilities of students' argument writing, which is critical to supporting students' writing development in AWE systems. We demonstrated that reliable manual annotation can be achieved and that the RER scheme correlates in largely expected ways with a holistic assessment of the extent to which revisions  address the feedback provided. We conclude that this scheme has promise in guiding the development of an automated revision classification tool. 

Although the RER scheme was developed with a specific corpus and writing assignment,  we believe some of the categories (e.g., reasoning linked to claim and evidence) can easily be adapted to data we have from other revision tasks. 
With more data, we also plan to improve the current classification method with state-of-the-art machine learning models, and extend the classification for all categories.  

\section*{Acknowledgments}
The research reported here was supported, in whole or in part, by the Institute of Education Sciences, U.S. Department of Education, through Grant R305A160245 to the University of Pittsburgh. The opinions expressed are those of the authors and do not represent the views of the Institute or the U.S. Department of Education.

\bibliography{references}
\bibliographystyle{acl_natbib}

\end{document}